\documentclass[letterpaper]{article} 
\usepackage{aaai25}  
\usepackage{times}  
\usepackage{helvet}  
\usepackage{courier}  
\usepackage[hyphens]{url}  
\usepackage{graphicx} 
\usepackage{natbib}  
\usepackage{caption} 
\usepackage{algorithm}
\usepackage{algorithmic}

\usepackage{newfloat}
\usepackage{listings}
\DeclareCaptionStyle{ruled}{labelfont=normalfont,labelsep=colon,strut=off} 
\floatstyle{ruled}
\newfloat{listing}{tb}{lst}{}
\floatname{listing}{Listing}
\usepackage{tabularx}
\usepackage{array} 
\usepackage{xcolor}
\usepackage{dirtytalk}

\title{Adoption of Explainable Natural Language Processing:\\ Perspectives from Industry and Academia on Practices and Challenges}
\author{
    Mahdi Dhaini, Tobias M\"uller, Roksoliana Rabets, Gjergji Kasneci
}
\affiliations{
    Technical University of Munich\\ School of Computation, Information and Technology\\ Department of Computer Science\\ Munich, Germany

    \{mahdi.dhaini, roksoliana.rabets, gjergji.kasneci\}@tum.de, tobias.mueller15@sap.com
    
}

\usepackage{bibentry}

\begin{document}

\maketitle

\begin{abstract}
The field of explainable natural language processing (NLP) has grown rapidly in recent years. The growing opacity of complex models calls for transparency and explanations of their decisions, which is crucial to understand their reasoning and facilitate deployment, especially in high-stakes environments. Despite increasing attention given to explainable NLP, practitioners' perspectives regarding its practical adoption and effectiveness remain underexplored. This paper addresses this research gap by investigating practitioners' experiences with explainability methods, specifically focusing on their motivations for adopting such methods, the techniques employed, satisfaction levels, and the practical challenges encountered in real-world NLP applications. Through a qualitative interview-based study with industry practitioners and complementary interviews with academic researchers, we systematically analyze and compare their perspectives. Our findings reveal conceptual gaps, low satisfaction with current explainability methods, and highlight evaluation challenges. Our findings emphasize the need for clear definitions and user-centric frameworks for better adoption of explainable NLP in practice. 

\end{abstract}

%

\section{Introduction}

The rapid advancements in artificial intelligence (AI) have driven its integration into a wide array of applications such as legal \citep{Vladika_Meisenbacher_Preis_Klymenko_Matthes_2024, zhong-etal-2020-nlp_survey_legal}, healthcare \cite{huang-etal-2024-nlp_llm_survey},  and finance settings \cite{Du_survey_nlp_finance_2025}; however, as these systems become more complex, they often resemble \say{black boxes}, making it difficult to understand how specific decisions or predictions are made. This opacity poses significant challenges in trust, accountability, and ethical deployment, giving rise to the need for Explainable AI (XAI) \cite{Hassija_Chamola_Mahapatra_Singal_Goel_Huang_Scardapane_Spinelli_Mahmud_Hussain_2024}. XAI aims to provide human-understandable insights into the workings of AI models, enabling stakeholders, such as developers, regulators, and end-users, to interpret and trust AI-driven outcomes \cite{Barredo_Arrieta_Díaz-Rodríguez_DelSer_Bennetot_Tabik_Barbado_Garcia_Gil-Lopez_Molina_Benjamins_etal._2020}.
In recent years, explainability has gained increased attention due to emerging regulations such as the European Union’s AI Act and the General Data Protection Regulation (GDPR) \cite{Nisevic_Cuypers_De_Bruyne_2024}. The EU AI Act emphasizes transparency, accountability, and risk management in AI systems, particularly in high-risk applications. GDPR, on the other hand, provides individuals with the \say{right to explanation} for decisions made by automated systems. These regulatory frameworks underscore the need for technical XAI developments to ensure compliance, enhance user trust, and mitigate risks associated with opaque AI models \cite{Hamon_Junklewitz_Sanchez_Malgieri_De_Hert_2022}.
Within the broader landscape of XAI and the rising popularity of Large Language Models (LLMs), explainability in Natural Language Processing (NLP) represents a critical and evolving area of research \cite{Gurrapu_Kulkarni_Huang_Lourentzou_Batarseh_2023}. In addition to prominent LLM applications, NLP models are widely deployed in applications such as sentiment analysis, information retrieval, and conversational agents, often operating on sensitive textual data. The interpretability of these models is crucial, particularly when their decisions directly impact individuals or involve subjective language \cite{lyu-etal-2024-faithful-survey-nlp}.
Recent literature actively investigates explainable AI \cite{Penu_Boateng_Owusu_2021, Lukyanenko_Castellanos_Samuel_Tremblay_Maass_2021, Haque_Islam_Mikalef_2023, Aryal_Sarkar_2024}, including some studies on its application in the NLP domain. For example, \cite{Wambsganss_Engel_Fromm_2021} explored how Feature Engineering methods can be applied to improve explainability for NLP-based systems, \cite{Sebin_Taskin_Mehdiyev_2024} conducted a systematic literature review to analyze the intersection between LLMs and XAI, while further studies performed surveys on explainable NLP \cite{Danilevsky_Qian_Aharonov_Katsis_Kawas_Sen_2020, Gurrapu_Kulkarni_Huang_Lourentzou_Batarseh_2023, lyu-etal-2024-faithful-survey-nlp}. 

Despite the growing body of research on explainable NLP, limited attention has been given to practitioners' perspectives regarding its practical application. Practitioners' experiences with explainability methods in real-world NLP settings, including their motivations for adoption, actual practices, the challenges encountered, and satisfaction with existing techniques, remain underexplored. Addressing this gap is crucial, as insights from practitioners provide a realistic view of explainable NLP's current effectiveness and utility. 
This paper aims to address this gap and explore potential solutions to mitigate challenges hindering the adoption and full utilization of explainable NLP in practice.

Consequently, we pose the following Research Questions (RQs):
\begin{itemize}
    \item \textbf{RQ1}: What are the goals and motivations of applied explainable NLP, and which methods are commonly used by practitioners in real-world NLP use cases?
    \item \textbf{RQ2}: What are the challenges and solution concepts for the adoption of explainability techniques within NLP use cases in practice?
\end{itemize}

As our contributions, we answer these RQs by conducting an interview-based study with practitioners to gather their perspectives on explainability methods and systematically analyzing their responses. Additionally, we interviewed academic researchers to assess whether, and to what extent, there is agreement or divergence between the perspectives of industry and academia on applied explainable NLP.

The results of our study reveal a lack of consensus on explainability definitions and varied interpretations of explanation. Additionally, we recognized faithfulness and simplicity as the main requirements for a useful explanation, and observed low satisfaction levels with the currently used explainability methods (73\% of interviewees). Key challenges identified among the practitioners include evaluating the quality and effectiveness of explanations. The results also emphasize the need for new evaluation frameworks for explainability methods and the importance of having user-centric explanations tailored for stakeholder groups.
In addition, we highlight the implications of our findings for both research and practice.

In the following, we first provide the theoretical background to our study by introducing the notion and background of Explainable AI and relevant concepts of NLP (Section 2). Subsequently, we describe the process and rationale behind the applied research methodology, including the study design, data collection, and data analysis, and methodological limitations of the semi-structured interviews conducted (Section 3). This is followed by an overview of the resulting findings structured by the posed RQs (Section 4). Then we reflect on the research problem and research gaps with a thorough discussion of our findings and contributions (Section 5). Finally, we conclude our work and outline potential future research (Section 6).

\section{Theoretical Background}
\subsection{Black-box Models in NLP}
Natural language processing is a field of research and application that investigates how computers can understand and manipulate natural language text or speech to perform useful tasks \cite{Chowdhary_2020}. Advances in NLP in recent years have included the introduction of deep learning architectures like BERT \cite{Devlin_Chang_Lee_Toutanova_2019} and GPT \cite{Brown_Mann_Ryder_Subbiah_Kaplan_Dhariwal_Neelakantan_Shyam_Sastry_Askell_et_al._2020}, leading up to the development of very large and complex models known as LLMs like ChatGPT (OpenAI, 2023). Modern NLP models are characterized by their high performance on several tasks and lack of transparency \cite{Danilevsky_Qian_Aharonov_Katsis_Kawas_Sen_2020}; these models are considered black boxes and provide no explanations for their decisions or decision-making processes.

\subsection{Explainable NLP}
The lack of transparency in deep learning models motivates the field of explainable AI, which aims to bridge the gap between model complexity and explainability \cite{Adadi_Berrada_2018}. Explainable NLP refers to the application of explainability techniques and methods to elucidate the decision-making processes of models used in NLP-based systems \cite{Danilevsky_Qian_Aharonov_Katsis_Kawas_Sen_2020}. The need for explainability in NLP is even more evident in high-stakes use cases, such as medical and legal domains, where understanding the model's decisions is of utmost importance for trusting its outputs.

Popular explainability methods applied in the NLP literature include feature importance methods such as SHAP \cite{Lundberg_Lee_2017}, LIME \cite{Ribeiro_Singh_Guestrin_2016}, and gradient-based methods \cite{gradient_survey}. Both SHAP and LIME are perturbation-based explanation methods where, for a given textual input, these methods determine the contribution of individual words to the model's final prediction by analysing changes in the prediction when specific words are perturbed and thus highlighting the importance of each word in influencing the final prediction \cite{Ivanovs_Kadikis_Ozols_2021}. In addition to perturbation-based methods, gradient-based methods use the model's output with respect to the input features to identify the contribution of each word to the final prediction \cite{Ivanovs_Kadikis_Ozols_2021}.

With the increasing focus on textual data and the widespread adoption of LLMs, it has become increasingly important to address \textbf{explainability in the context of NLP}. This is particularly critical in high-stakes domains such as law, finance, and healthcare, where understanding model behavior is essential. Research is therefore needed to examine the use of explainability methods across diverse NLP use cases, especially those involving sensitive or consequential decision-making. In addition, textual datasets pose unique challenges for explainability that differ significantly from other data types, such as tabular data \cite{dhaini2025genderbiasexplainabilityinvestigating}, which was the main focus of prior research and thus underscores the need for explainability studies tailored specifically to NLP. Findings from general ML domains cannot be directly transferred to NLP settings, further motivating the importance of NLP-specific investigation. Despite the growing body of work on explainable NLP in the literature, relatively little attention has been given to studying its state in practice. Previous explainable NLP studies \cite{Habiba_ML_practitioners_perceive_2024, krishna_disagreement_2024}
lack a thorough understanding of applied explainable AI in NLP. Investigating this gap serves as the primary motivation for this paper. 

To the best of our knowledge, this is the first study to focus on the practitioners' perspectives on applied explainable NLP and investigate its current state in practice through a thematic analysis of interviews with practitioners and researchers from various domains.

\section{Research Methodology}
To investigate the status quo of explainable NLP in practice, we followed a qualitative research approach and drew on the experiences of experts by conducting semi-structured interviews. Due to the novelty of the topic and to examine the potential gap between research and practice, we first conducted an initial literature review \cite{Danilevsky_Qian_Aharonov_Katsis_Kawas_Sen_2020, balkir-etal-2022-challenges, Freiesleben_König_2023, longo-2023, Zhao_Chen_Yang_Liu_Deng_Cai_Wang_Yin_Du_2024} to build our knowledge base and as the foundation for the questionnaire. Subsequently, we interviewed experts from academia and industry to determine whether a gap exists between research and practice in the field of explainable NLP. The following sections describe the study design, data collection, and data analysis of our research.

\subsection{Study Design}

\begin{table*}[ht]
\centering
\small
\begin{tabular}{clllllc}
\hline
\textbf{ID} & \textbf{Position} & \textbf{Organization} & \textbf{Size} & \textbf{Industry/Academia} & \textbf{Country} & \textbf{Experience (years)} \\
\hline
I1 & Research Associate & University & - & Academia & India & 2 \\
I2 & Research Associate & University & - & Academia & Estonia & 2 \\
I3 & ML \& NLP Engineer & Software Company I & Medium & Industry & US & 5 \\
I4 & Data Scientist & Applied AI Consultancy & Small & Industry & Ireland & 2 \\
I5 & Research Associate & University & - & Academia & Germany & 6 \\
I6 & Data Scientist & Software Company II & Medium & Industry & US & 3 \\
I7 & Data Scientist & Software Company III & Medium & Industry & US & 4 \\
I8 & Postdoctoral Researcher & University & - & Academia & Germany & 6 \\
I9 & Data Scientist & R\&D Lab & Start-Up & Industry & Bangladesh & 3.5 \\
I10 & Staff Data Scientist & Order and Delivery Platform & Large & Industry & Germany & 10 \\
I11 & Researcher & University & - & Academia & Netherlands & 6 \\
I12 & Data Scientist & Software Company IV & Small & Industry & Germany & 5.5 \\
I13 & ML Engineer & Cybersecurity Company & Small & Industry & Pakistan & 2 \\
I14 & AI/ML Model Steward & Financial Services & Large & Industry & Canada & 3 \\
I15 & Data Scientist & Software Company V & Large & Industry & Germany & 6 \\
\hline
\end{tabular}
\label{tab:interviewee_demographics}
\caption{Overview of Interviewee Demographics}
\end{table*}

To capture insights and experiences from NLP practitioners and researchers, we utilized semi-structured interviews, which provided a structured yet flexible approach to inquiry. To identify potentially relevant interviewees working on explainable NLP either in research or business, we contacted pre-saved contacts of prior research and referrals. We assume expertise when the individual has 2 years or more of experience in the field of explainable NLP. Additionally, we researched companies and individuals that specialize in NLP through top search results (e.g., LinkedIn). Hereby, we reviewed the background of each potential candidate and only considered individuals with experience in NLP and explainability or interpretability in further steps. Then, we contacted the identified experts with relevant topic-related knowledge and sent invitations through email or direct message. After a positive response, we scheduled the interview.

We contacted 94 experts, of whom 14 expressed interest in participating in an interview. Through a referral during the interview process, we were able to conduct a total of 15 interviews. Table 1 provides a codified overview of our resulting sample, including relevant information such as their position, organization, and years of experience in the field of NLP. From here on, we will reference the experts by their corresponding unique participant identifiers (ID). The interviewee demographics include 5 researchers and 10 industry experts from 9 countries, 10 different companies (and 5 universities) of varying sizes with a broad spectrum of backgrounds, job profiles, and experience. 

Thereby, we aimed to include a large variety of voices (Myers and Newman 2007) and to cover multiple viewpoints. For the interviews, we followed the guidelines proposed by Myers and Newman (2007). Prior to the scheduled interviews, we briefed the interviewees on the research purpose, content, and structure to allow for follow-up questions. Additionally, we sent the questionnaire beforehand. All interviews were recorded, transcribed, and coded. The resulting insights were iteratively compared to findings from previous interviews. Finally, we concluded our interview study after reaching theoretical saturation. Since the interviewees only allowed the findings to be published anonymously, we did not include the full transcripts.


\subsection{Data Collection}

The semi-structured interviews were conducted via Zoom in the time span between December 2023 and September 2024. To ensure observer triangulation \cite{Runeson_Höst_2009}, each session included two researchers. First, participants were informed about the recording and future use of the interview transcripts, including data anonymization protocols. Additionally, we reminded the interviewees of the research goals and the interview structure to ensure clarity. We kept a consistent format and outline for each interview, and the questions remained unchanged. Nevertheless, the flexible nature of semi-structured interviews allowed for minor adjustments in the order of questions or wording in response to the flow of the conversation. The interview questions were derived from themes identified in the preliminary literature review and comprised five thematic sections:

\textit{(1) Background and Common Basis:} The first section aimed to gather foundational data on the participants, including their professional background, and experience in NLP and explainability/interpretability. Moreover, we asked the participants to define and differentiate between “explainability” and “interpretability” in the context of NLP. These insights allowed us to contextually interpret their response and ensure that the insights are analyzed with consideration of the corresponding background and individual understanding of the core concepts.

\textit{(2) Explainable NLP:} This was followed by a section on the motivation for adopting explainability techniques, focusing on their primary objectives in implementing these methods within their work. Moreover, we asked the interviewees to define a “good explanation” of the NLP system, including potential qualities and criteria they value in explainable outputs. Aiming towards RQ1, we thereby gathered information on the interviewees’ motivation for implementing explainability and their requirements for useful explanations.  

\textit{(3) Explainability Methods:} This section also aims to answer RQ1 by investigating which explainability methods are used by participants in practice to get explanations for the models’ decisions. We first asked the interviewees about the explainability methods they use. Then, we grouped the interviewees' answers based on categories that we had already derived from the literature. We then assessed the level of satisfaction with the identified methods based on the interviewees' expectations and objectives. 

\textit{(4) Challenges and Solution concepts:} To answer RQ2, we asked the interviewees to reflect on the challenges and solution concepts of working with explainable models in real-world settings.

\textit{(5) Closing:} Finally, we provided the opportunity to share any additional insights that may not have been addressed by the structured questions. Hereby, we also asked if the experts would be open to follow-up contact or could recommend other potential participants.

\subsection{Data Analysis}

The recorded interviews were transcribed with the help of \textit{otter.ai} and systematically coded following the Reflexive Thematic Analysis (RTA) process as proposed by Braun and Clarke (2019), where the researcher has an active role in thematic analysis and "the researcher’s reflective and thoughtful engagement with their data" \cite{Braun_Clarke_2019}.

More specifically, the process is structured into six steps and was conducted as follows:
\begin{enumerate}
    \item First, the researchers familiarized themselves with the data. The initial insights from each interview were noted and contextualized within the broader dataset. Each expert was assigned a unique identifier, and any potentially sensitive information was removed to maintain participant anonymity.
    \item The entire dataset was coded to capture meaningful features. Codes were assigned to important features relevant to the RQs. New codes were developed whenever findings could not be accommodated within existing categories, prompting a re-coding of earlier data. This iterative coding process involved multiple rounds to refine the final set of codes.
    \item Initial themes were generated by identifying patterns and clustering the codes into broader themes.
    \item The themes were reviewed against the coded data and refined to ensure that the themes accurately reflected the data in relation to the RQs.
    \item Each theme was defined and named on the basis of its core concept. Thereby, the researchers developed a detailed analysis including the scope, focus, and naming of each theme.
    \item Lastly, the interviews and thematic findings were summarized and analyzed in relation to existing literature.
\end{enumerate}

Emerging discrepancies throughout the RTA process were discussed among the researchers and resolved by consensus. The resulting themes are described in detail in section 4.

\subsection{Methodological Limitation}
Limitations of our methodology include the relatively modest sample size. Although the sample size remains a reasonable number compared to similar studies in the literature, the resulting insights do not allow generalization to uninvestigated domains.
The practitioners’ perspectives in our study are based on nine participants out of 15, as the remaining six are from research institutions. Still, this sample size of practitioners remains comparable to that of similar studies \citep{warren_2025_fact_checkers}, and insights from the six research participants were incorporated as supplementary input for comparison.
Also, we observed that no new themes were emerging in the later interviews, indicating that thematic saturation had been reached. All interviews were conducted in English; as shown in Table 1, our sample was diverse and included participants from several non-native English-speaking countries. Conducting interviews in additional languages might have enabled a larger and potentially more inclusive sample.

We therefore encourage practitioners, researchers, and other interested readers to use our work as a foundation to complement and refine our results, potentially extending this research with their own insights and with follow-up studies to verify our presented findings and results.  

Due to the reflexive nature of the RTA approach and its emphasis on the researcher's active and thoughtful engagement with both the data and the analytic process \cite{Braun_Clarke_2019}, it is important to acknowledge that the interpretations produced are inherently shaped by the individual researcher. Consequently, while it is possible for another researcher to generate similar codes or themes, such reproducibility should not be expected \cite{Braun_Clarke_2019}. However, this is not considered a main limitation of the methodology but more of a characteristic of the RTA framework as variability in code and their development is expected and accepted \cite{Braun_Clarke_2019}. Although we carefully designed the interviews to ensure a consistent and neutral process and made efforts to maintain reflexivity in our analysis, the results from the interviews may still be subject to the inherent subjectivity associated with self-reported data \cite{donaldson2002understanding}.

A note on reporting quantitative results in thematic analysis: in this thematic analysis study, counts and percentages are reported to indicate how frequently certain themes were mentioned among participants, thereby enhancing transparency and providing a sense of the distribution of responses within this sample. Given the modest sample size (n = 15), these percentages should be interpreted cautiously and not regarded as measures of statistical significance or generalizability \citep{Braun_Clarke_2019, Sandelowski_2001_numbers_qualitative_research, Maxwell_2010_numbers_qualitative_research}.

\section{Results}
\subsection{Explainability and Interpretability}

A standard or shared definition of terms is essential for a common understanding within a community and to advance the field. In the XAI community and research, the two main terms used are explainability and interpretability; however, we don’t have a clear distinction between the two terms. Thus, we first investigate how these terms are defined and used and present results accordingly. In this section, we present and examine the results of our interview study, where we asked interviewees whether they distinguish the two terms, the definition and usage of the terms, and the need for having strong definitions. Around 46\%  of the participants (7 out of 15) agreed that these terms can be used interchangeably, while others indicated a clear distinction between the two terms such that they can’t be used interchangeably.

\paragraph{Definition of Explainability}
This term is seen as a broader term compared to interpretability, even for participants who agree on the interchangeable use of both terms. Around 66\% of participants (10 out of 15) indicate that explainability refers to the explanation of a model’s output and decision “regardless of the model’s inner workings and process” (I2). Based on the interviewees’ responses, we identified the following common features and themes when defining or referring to explainability: (1) Referring to a specific or particular prediction of a model, (2) understanding why a prediction is made by a model and (3) referring to explaining the model to the end user.

\paragraph{Definition of Interpretability}
This term is seen as a technical term compared to explainability, where 10 out of 15 interviews agreed that it refers to understanding the model itself, including its internals, including its inner workings, regardless of a specific output. This includes understanding “how model’s weights are updated” (I13). Common features and themes that are associated with defining interpretability include: 
(1) referring to the understanding of any process within the model's inner workings, and (2) referring to the model properties. 

Hence, the majority of participants agreed on the importance of having common definitions for such terms for the purposes of a better understanding of the research field. This would be especially helpful for newcomers and for the wider and easier adoption of explainability in practice “if we are building tools and models, which are consumer-facing or can have an impact on actual lives of the people, then having actual definition certainly helps in making sure we are making it the right way” (I10)

Results from our initial literature review show that, when referring to explainability and interpretability, (1) no distinction is made, and they have the same definition, (2) the two terms are distinct, and (3) no clear consensus on the definition of each term when a distinction is made. This lack of consensus on defining and using these two terms can have substantial effects on implementing explainability in practice. 


For the rest of this paper and to present results simply, we will use the term explainability instead of interchangeably using both terms.  

\subsection{Motivation and Goals}
Understanding the motivations and goals of researchers and practitioners in using explainable NLP is essential as it allows aligning the advancements and developments in this research field with the actual needs and expectations of practitioners, which increases the likelihood of adopting explainability methods and techniques in practical NLP use cases and applications. In addition, identifying such motivations and goals can help highlight challenges researchers and practitioners face. In this section, we present the interview results, in which participants were asked about their primary motivations and objectives for working with or utilizing explainable NLP methods and techniques. Additionally, we explore whether they consider explainability a main requirement in their use cases and projects. We present the results for these points in Table 2. We also follow the categorization framework proposed by \cite{Adadi_Berrada_2018}, which classifies motivation into the following categories: Explain to \textbf{justify}, explain to \textbf{control}, explain to \textbf{improve}, and explain to \textbf{discover}.  


\begin{table*}[ht]
\centering
\small
\begin{tabular}{p{4.5cm}p{9cm}}
\hline
\textbf{Category} & \textbf{Motivations (Participants)} \\
\hline
Explain to justify &
Achieve transparency and user trust (I1, I5), \newline
Have acceptance and trust of stakeholders (I2), \newline
Compliance with regulations and lack of understanding (I6), \newline
Transparency and confidence in model’s outputs (I7), \newline
Lack of users trust and understanding (I10, I14), \newline
Ease the creating of policies and moderation rules (I11), \newline
Lack of understanding and knowledge extraction (I12) \\
\hline
Explain to control &
Debug model (I3), \newline
Accountability assurance (I8) \\
\hline
Explain to control and to justify &
Debug model and understand model’s decisions (I4), \newline
Debug model and lack of understanding (I13) \\
\hline
Explain to justify and control &
Lack of users trust, debug model, impact of models (I9) \\
\hline
Explain to justify and to improve &
Satisfy user requirement to understand model predictions, \newline
continuously improve model performance (I15) \\
\hline
\end{tabular}
\caption{Grouped results on the motivation and goals of using explainability in NLP by interviewees.}
\label{tab:motivation_goals_grouped}
\end{table*}

Most motivation results fall under the \textit{explain to justify} category, where the motivations include having justifications and reasons for outcomes and results of the NLP model to build trust and comply with legislation \cite{Adadi_Berrada_2018}. Common motivations and goals of participants for using explainable NLP include having explanations on why the model makes certain predictions to achieve users' trust in the model. As referred by some interviewees, some clients, such as doctors and patients in a medical setting, “fear” trusting and implementing NLP black-box models that provide no explanations for their predictions; that’s why it is important “to remove the blockage of the AI implementations and reassure the clients by providing models that can explain their decisions” (I4). Another common motivation is having explanations for the model’s unexpected decisions to debug the model to improve its performance in different settings and tasks, such as machine translation (I6-I7) and in the legal domain (I14).


\paragraph{Explainability as a requirement} 
After understanding the experts’ motivation for incorporating explainability, we assessed whether it had been considered a requirement or a feature in NLP projects they worked on. We asked participants if explainability has been considered in previous or ongoing projects as a requirement or additional feature, or not been considered. In this context, we differentiate between a requirement essential to the system and an added feature that enhances users' experience and trust in the system. 

For 7 out of 15 experts (46.6\%), explainability was a main requirement in their works, while it was an additional feature for 5 experts (33.3\%), and 3 experts (20\%) mentioned that it was not considered in their works as a main requirement or an additional feature. Explainability was indicated as a main requirement for interviewees working on projects in high-stake domains like finance (I14, I15), medicine (I10) and resume matching (I3) where in a medical data usage-related project, “it was a definite requirement to have explainability built into the solution” (I10) where users wanted to “to see a machine learning model on the basis of what features they make decisions” (I10). 

\paragraph{Useful explanations}
As not every explanation can be useful, identifying the properties and features that make an explanation useful helps improve explanation methods to provide explanations that satisfy these aspects.  We asked the experts about the features they consider important in a provided explanation, so it can be a useful explanation. The characteristics and features of a useful explanation are related to the motivation and goal of the experts to use explainability methods. As presented earlier, the most common motivation for having explainable NLP is explaining and justifying, so many of the interviewees' answers were related to having explanations with specific features and characteristics that facilitate model debugging and a better understanding of the model's decisions. 

Although the domain in which these explanations are required may impact the definition of a helpful explanation, some features were standard across different domains and shared between the experts’ answers. Based on the interviewee's answers, an explanation is useful when: (1) it \textbf{faithfully} reflects the model’s reasoning and output-making process “so that it actually reflects what the model does” (I5), (2) is \textbf{straightforward and easily understandable} within the context of the model, provides insights into the decision-making process “like the most important features for making a decision” (I9), and provides insights that ease debugging the model such as “pointing out data samples that influence wrong decisions” (I2).  

\subsection{Explainability Methods}

Given the large number of explainability methods introduced in the explainable NLP community \cite{Danilevsky_Qian_Aharonov_Katsis_Kawas_Sen_2020, Zini_Awad_2022}, there is a lack of insight into which methods are being used by practitioners. To better understand which explainability methods are most used, we asked experts about the methods and techniques they have used in their NLP projects. 
SHAP-based methods were the most mentioned methods by interviewees (8 out of 15), followed by LIME (7 out of 15). These methods were popular among the experts as each indicated that they used either or both methods in their use cases. The third-most used methods were gradient-based ones, such as Integrated Gradients \cite{Sundararajan_Taly_Yan_2017} (6 out of 15). Experts apply these methods in different use cases, such as hate-speech detection (I2), depression detection (I1), resume-to-job matching (I3), and misinformation detection (I11). For instance, some experts shared how these methods can be helpful in insurance and financial projects, as it "was helpful to understand why the model was, for example, accepting or rejecting a claim."(I10)

In addition to the methods above, other methods indicated by the experts include attention-based methods like saliency scores (I5), explainable word embeddings (I2), and the use of decision trees as white box models for text classification (I14), which is inherently explainable where no additional operation is needed to explain the model decisions \cite{Danilevsky_Qian_Aharonov_Katsis_Kawas_Sen_2020}. Notably, experts indicated that these methods were used in addition to using SHAP and/or LIME, mostly to get different types of explanations or to compare the explanations.
We present these results in Table 3, grouping the explanation methods used by interviewees according to their categories and providing the count of methods employed within each category.

\begin{table}[ht]
\centering
\small

\begin{tabular}{lc}
\hline
\textbf{Category} & \textbf{\# Experts (Occurrences)} \\
\hline
Perturbation-based methods & 15 (100\%) \\
Gradient-based methods     & 6 (40\%) \\
Other                      & 3 (20\%) \\
\hline
\end{tabular}
\label{tab:explanation_methods}
\caption{Categories of explanation methods used by experts.}
\end{table}

The most commonly used explainability methods among experts were perturbation-based and gradient-based techniques. These approaches fall under the category of feature attribution (also referred to as feature importance) methods, which emphasize how individual features contribute to the model's predictions \cite{Zhao_Chen_Yang_Liu_Deng_Cai_Wang_Yin_Du_2024}. This suggests that experts prioritize understanding the contribution of each feature to the model's output, enabling them to identify the most relevant features influencing the predictions.


\paragraph{Satisfaction with explainability approaches} 
To assess the level of satisfaction with the existing explainability methods they use, we asked interviewees if they believe they are sufficient for them based on their motivation and goals. Table 4 presents these results, where only 4 out of 15 experts declare that the explainability methods they use satisfy their goals of using these methods. 

\begin{table}[ht]
\centering
\small

\begin{tabular}{lc}
\hline
\textbf{Level of Satisfaction} & \textbf{\# Experts (Occurrences)} \\
\hline
Sufficient      & 4 (26.67\%) \\
Not Sufficient  & 11 (73.33\%) \\
\hline
\end{tabular}
\label{tab:expert_satisfaction}
\caption{Level of satisfaction of experts with explainability methods}
\end{table}

\subsection{Challenges and Solution Concepts for Adopting Explainable NLP}

To identify the challenges of adopting Explainable NLP, we asked the interviewees to reflect on the challenges and solution concepts of working with explainable models in real-world settings. For this section, we mainly drew from the experiences of the experts from the industry.

\subsubsection{Challenges for Adopting Explainable NLP} After coding and summarizing the interviews, \textbf{we
identified} ten challenges and grouped them into six categories as follows:

\paragraph{(1) Conceptual and Theoretical:}
These challenges involve difficulties in defining clear frameworks, principles, or models for interpreting and justifying the decisions and outputs of natural language processing systems.

\textit{Lack of formalism:} The lack of formalism in explainable NLP makes adopting explainability on NLP use cases more challenging. This is reflected by the ambiguous terminology, lack of consensus on standards and clear definitions, and lack of established guidelines and best practices in implementing explainability methods.

\textit{Disparities in understanding explanations:}  One main challenge described by experts is to ensure that explanations are meaningful and easy to understand for different stakeholders, where an explanation that is meaningful to them might not be understandable and unclear to other counterparts, like their project partners or clients.  

\textit{Interpretation of regulations:} This challenge highlights the difficulties faced by experts from the industry in complying with new regulations, such as the EU AI Act, due to the lack of clarity on identifying and interpreting the associated requirements.

\textbf{(2) Computational and resource:} These challenges include challenges regarding the high computational costs, scalability issues, and the need for extensive data and infrastructure to create and evaluate interpretable models.

\textit{Computational and memory requirements:} A challenge that experts face is the high computational and memory requirements for getting explanations using the explainability methods and implementations they work with. 

\textit{Reduced real-time usability:} This challenge is closely linked to the previous one, as a lack of resources often results in high latency when generating explanations for model decisions, which can significantly hinder the usability of an NLP-based system.

\textit{Explainability for larger models:} The introduction of large language models, such as ChatGPT, presents additional challenges in explaining their outputs. Applying the explainability methods commonly used by experts becomes infeasible due to the substantial size and complexity of these models.

\textbf{(3) Performance trade-offs:} \textit{The trade-off between the model’s explainability and performance} was a common challenge among experts. Having explainable by-design models often leads to a dip in performance. Many experts agreed that this challenge becomes less critical for low-stakes use cases, where having high performance is the main requirement. However, this challenge becomes more significant in high-stakes domains, like medical and legal applications, where both high performance and explainability are required. 

\textbf{(4) Evaluation and validation:} A common challenge for experts was \textit{evaluating the quality and effectiveness of explanations}. Some experts highlighted the lack of consensus on standardized evaluation metrics and methodologies, while others emphasized the difficulties they encounter when comparing explanations from different explainability methods due to the lack of benchmarks. Additionally, the lack of ground truth in the form of human-annotated explanations, which are often unavailable or costly to obtain, further deepens this problem. 

\textbf{(5) Tooling and implementation:} Some experts highlighted challenges in the \textit{usability of certain explainability methods} and their various implementations due to the lack of maintenance of the respective libraries and packages. This lack of maintenance complicates the application of these methods, increasing the effort required to implement them.

\textbf{(6) Language and domain:} As described by experts, obtaining high-quality and useful \textit{explanations is more challenging when working with models on non-English datasets} than when working with English ones. As explainability methods primarily aim to explain the decisions and decision-making processes of NLP models, this difficulty stems from existing challenges in non-English NLP, such as the scarcity of large and high-quality datasets relative to those available for English.
Furthermore, another challenge is the lack of domain knowledge when experts use explainability methods to generate explanations in domains outside their expertise. This issue was highlighted by the example of an NLP engineer attempting to interpret explanations for a model's decisions on medical texts. The absence of specific domain knowledge means that the explanations users may not fully understand the explanations of these methods, rendering them less useful and potentially leading to their disregard.

\subsubsection{Solution Concepts for Adopting Explainable NLP}
Similarly, we gathered experiences on the proposed solution concepts from the industry experts. Thereby, we identified seven solution concepts to overcome challenges for adopting explainable NLP and present them as follows:

\textbf{(1) Maintaining and extending explainable NLP frameworks:} The availability of well-maintained repositories and libraries for explainability methods facilitates the efficient and easy application of these methods. Additionally, extending existing frameworks and tools, which currently provide explainability methods only for a fixed set of classification tasks, to include explanation methods for other tasks, such as text generation, can further enhance the implementation and usage of explainability methods across different tasks, thereby increasing the adoption of explainability in NLP.

\textbf{(2) Benchmarks for evaluating explanations:} One solution concept in overcoming the evaluation challenge of explanations is the construction of standardized tasks- and domain-specific benchmarks that researchers and practitioners can use to evaluate the quality of explanations both quantitatively and qualitatively. 

\textbf{(3) Extending datasets with explanations:}  In addition, some experts refer to the importance of having explainable datasets \cite{wiegreffe2021teach}  where datasets are augmented by human-annotated rationales that serve as ground truth for evaluating explanations. Such datasets facilitate the evaluation of explanations generated by explainability methods. 

\textbf{(4) Requirements for regulation compliance:} Another factor contributing to the successful adoption of explainability is the availability of precise specifications of explainability requirements for NLP-based systems that must be satisfied to comply with regulations, especially in high-stakes domains. This set of requirements enables researchers and practitioners to assess whether existing explainability methods can satisfy these requirements or, if not, to design and develop new approaches that help align NLP-based systems with regulations and build regulation-compliant systems.

\textbf{(5) Open-source explainable NLP platform:} A key solution concept identified by experts is the development of an open-source platform dedicated explicitly to explainability, similar to an explainability-oriented version of Hugging Face \footnote{https://huggingface.co/}. Such a platform would enable the explainable NLP community to collaborate on all aspects of explainability frameworks, methods, evaluation metrics, explainable datasets, and applications. This platform could serve as a common ground for practitioners in explainable NLP, fostering easier adoption and implementation of explainability methods in both academia and industry.

\textbf{(6) Efficient explainability methods:} Several experts identified the development of memory- and computationally-efficient explainability methods as a significant solution concept for the continued adoption of explainability in NLP. For wider adoption and applicability, explainability methods should consume less memory and require less computational time, thereby providing explanations more quickly than existing methods. This is especially important given the ever-increasing size of language models.

\textbf{(7) User-specific explanations:} Another solution concept highlighted by experts is the provision of explanations tailored to the specific stakeholders involved. Explanations provided by explainability methods should not be considered universal solutions that benefit all stakeholders in a use case; instead, they should be customized based on the expectations and needs of each type of user. Additionally, explanations should consider the domain knowledge and technical expertise of the different users involved to ensure they are helpful.

\section{Discussion}
In this section, we elaborate on our contributions in relation to the research questions guiding our study, analyze and discuss our findings, and outline potential directions for future work.




\paragraph{Conceptual misalignment and shared motivations. (RQ1)}
Results from the semi-structured interviews revealed a \textbf{lack of consensus on explainability-related definitions}. Interestingly, a comparison between responses from practitioners and researchers showed that practitioners tended to use more consistent and simplified terminology, whereas researchers exhibited greater variability. This divergence highlights a conceptual gap in how explainability is understood and defined across stakeholder groups. Such disparity may hinder the effective adoption of explainability techniques in practice, and was echoed in the perceived lack of formalism noted by several experts. These findings underscore the need for the XAI community to develop clear, shared, and operational definitions that can be embraced by both researchers and practitioners to support the broader implementation of explainability in NLP.

Common motivations and goals reported by participants for using explainable NLP include the need to understand why a model makes certain predictions, particularly to foster user trust, where results addressing RQ1 indicate that the dominant motivation for adopting explainability in NLP is to explain and justify model decisions. \textbf{The alignment in motivation between researchers and practitioners} suggests that both groups view explainability as essential for fostering trust and meeting regulatory requirements. This shared perspective was further reflected in the high number of participants who described explainability as a core requirement or valuable feature in their NLP projects. Their responses emphasize that end-users often seek a clear understanding of model outputs, reinforcing the role of explainability in building trust and supporting the responsible deployment of NLP systems.

\paragraph{Popular explainability methods in NLP practice and anticipated transitions (RQ1)}
Results on the most used explainability methods in practice showed that SHAP and LIME were the most used by experts in their NLP projects. Not surprisingly, these feature attribution methods are widely used, given their extensive application in the literature. These results highlight alignment in the most used methods between researchers and practitioners and demonstrate that these methods remain useful and applicable, particularly in practice, as all industry experts we interviewed employ them. 

These results align with results from related work \cite{Habiba_ML_practitioners_perceive_2024}, as LIME and SHAP, which have been widely used in the explainable NLP literature. \textbf{However, further work is needed to ensure that explanations generated by such methods satisfy users' expectations and motivations.} This alignment could also be influenced by the fact that most of the use cases reported by the experts involve classification tasks (e.g., I1, I2, I10, I11, I13) where methods like SHAP and LIME are most suited to. On the other hand, practitioners should be aware of the limitations of these explanation methods \cite{Freiesleben_König_2023}, particularly regarding their fairness \cite{dhaini2025genderbiasexplainabilityinvestigating}.  

The use of explainability methods in practice, along with the fact that feature attribution techniques like SHAP and LIME are not directly applicable to LLMs due to their substantial computational and memory demands \cite{Deiseroth_Deb_Weinbach_Brack_Schramowski_Kersting_2024}, suggests that \textbf{LLMs are not necessary for every NLP task.} In many cases, smaller models tailored to specific applications may suffice, allowing for the application of more lightweight and established explainability methods. However, for tasks that do require large models, practitioners may consider alternative explanation techniques designed for LLMs \cite{Deiseroth_Deb_Weinbach_Brack_Schramowski_Kersting_2024, Leemann_Fastowski_Pfeiffer_Kasneci}. As part of future work, \textbf{we plan to study how practitioners engage with LLMs in their NLP workflows.} We anticipate a reduced reliance on traditional post-hoc methods such as SHAP and LIME, and a growing adoption of explanation approaches tailored to LLMs \cite{Zhao_Chen_Yang_Liu_Deng_Cai_Wang_Yin_Du_2024}. Nonetheless, qualitative research remains essential to understand how practitioners perceive these emerging methods and whether they consider them useful and actionable in practice.

\paragraph{Desired properties for useful explanations (RQ1)}
Our results indicate that among various explanation properties, \textit{faithfulness and simplicity} were the most valued by practitioners, highlighting their critical role in practical applications of explainable NLP. \textbf{Faithfulness}, or fidelity, refers to how accurately an explanation reflects the underlying model’s actual decision-making process \cite{Danilevsky_Qian_Aharonov_Katsis_Kawas_Sen_2020}. It is regarded as one of the most important desiderata of model explanations in explainable NLP \cite{lyu-etal-2024-faithful-survey-nlp}. The alignment between practitioners and researchers regarding the importance of faithfulness is encouraging; however, recent research highlights inconsistencies and limitations in existing metrics for assessing explanation faithfulness \cite{zhao_aopc_unfaithfulness_2022, zhao-aletras-2023-incorporating_fiathfulness_metrics}. These discrepancies can hinder practitioners from adopting explainability methods, as they rely heavily on accurate metrics to evaluate faithfulness. \textbf{Simplicity}, also referred to as sparsity or comprehensibility, measures how easily explanations can be understood by users \cite{Danilevsky_Qian_Aharonov_Katsis_Kawas_Sen_2020}. While multiple metrics exist to evaluate simplicity, the importance of this property varies significantly among stakeholders. For instance, laypeople typically prioritize simplicity more than developers, who may emphasize faithfulness or other technical attributes. Although simplicity generally carries less criticality compared to faithfulness, it was a main sought desideratum of model explanations in the interviews. Its prominence as a practitioner requirement underscores the need to address it explicitly. Consequently, future research on explainability methods should rigorously evaluate both faithfulness and simplicity of explanations, complemented by extensive human-centered studies, to ensure alignment with practitioners' expectations and requirements, thereby enhancing practical usability in NLP applications.

\paragraph{Low satisfaction levels with explainability methods (RQ1 \& RQ2)}
Our results indicate low satisfaction with the explainability methods currently used by experts in NLP, highlighting the need to better align both new and existing methods with users' goals and expectations. Integrating human feedback into the design process is essential to ensure that these methods address stakeholder needs effectively. Furthermore, we emphasize the importance of incorporating human-centered evaluation in the benchmarking of new explainability techniques. As noted in prior work\cite{Nauta_Trienes_Pathak_Nguyen_Peters_Schmitt_Schlötterer_van_Keulen_Seifert_2023}, many proposed methods are introduced without any form of human evaluation. These findings underscore the necessity of developing domain- or stakeholder-specific explanation approaches. By first identifying the expectations, needs, and preferred characteristics of explanations for particular user groups, methods can be designed or adapted to more meaningfully support real-world decision-making.



\paragraph{Need for tailored and user-centric explanations (RQ2)}
To answer RQ2, we identified ten practical and conceptual challenges broken down into six categories that restrict the adoption and broader application of explainable NLP. We then identified seven solution concepts to overcome these challenges. The identified solution concepts pave the way for efforts to achieve them, facilitating the easier adoption and application of explainability methods.
An important insight drawn from our results is the need for tailored explanations where the same explanations cannot be expected to satisfy the needs and expectations of all stakeholders involved. Different stakeholders have varying motivations for requiring explainability and, consequently, different expectations for what constitutes useful explanations. This is reflected in the key challenge of disparities in understanding explanations when adopting explainability in NLP, and the solution concept of having user-specific explanations that we identified. Therefore, further work should focus on understanding the expectations and needs of different groups of stakeholders as a prerequisite to designing explanations based on the specific requirements of each group. These results also complement some proposed solutions in the related literature where \cite{longo-2023} propose supporting human-centeredness of explanations by creating human-understandable explanations and facilitating explainability with concept-based explanations.  

\paragraph{Explanation evaluation continues to be problematic (RQ2)}
One key challenge we identified, common among experts interviewed, was evaluating the quality and effectiveness of explanations. We also identified constructing benchmarks for evaluating explanations and extending datasets with explanations as solution concepts to overcome this challenge. These findings align with and complement existing literature \cite{longo-2023}, which identifies the lack of standardized metrics and methods as a key challenge in evaluating explainability systems. The literature also emphasizes the need to develop new evaluation frameworks for explainability methods and to address the limitations of and the overemphasis on human-based studies \cite{longo-2023, Freiesleben_König_2023}.

\paragraph{Implications for future research and practice (RQ2)}
The set of challenges and solution concepts we identified provides a foundation for actionable research, encouraging future studies to prioritize user-centric and stakeholder-driven explanation approaches. For \textbf{researchers}, our findings underscore the substantial room for improvement in explainable NLP, particularly in adapting existing methods and embracing user-centered principles in the design and development of new explainability techniques to meet stakeholder expectations and needs. For \textbf{practitioners and industry}, these findings can inform more effective adoption and integration of explainability tools, helping to reduce barriers to practical deployment and fostering broader application in real-world NLP systems.

\textbf{For regulators}, our findings indicate that current explainability methods, particularly traditional post-hoc techniques such as SHAP and LIME, although widely used by NLP practitioners, have not yet reached a level sufficient to meet practitioners' (and potentially laypeople’s) expectations and satisfaction. This limitation could diminish the value of employing these methods in NLP use cases, particularly in high-stakes domains where explainability is mandated by regulations, such as the EU AI Act. Consequently, these shortcomings might also hinder effective regulatory compliance, particularly if existing explainability methods fail to fulfill stakeholders' expectations and needs. For developing and updating regulatory guidelines, regulators should thus carefully consider the practical effectiveness of current and emerging explainability methods, including recent approaches such as mechanistic explainability in NLP \cite{mechanistic_Sarah_2024} that might be seen as tools for complying with regulations.




\section{Conclusion}
This paper presents the first qualitative study examining the state of explainable NLP in practice by focusing on practitioners’ experiences, which remain underexplored in the literature. Hereby, we focused on their perspectives on explainability methods in real-world settings, including their motivations for adoption, implementation practices, encountered challenges, and satisfaction with existing techniques. Following a qualitative research approach, we began with an initial literature review to build a foundational knowledge base and inform the development of our questionnaire. Subsequently, we conducted semi-structured interviews to capture the experiences of experts from both academia and industry. Our findings and results were presented and discussed, offering valuable insights into the current state of explainable NLP in practice. This work complements the current explainable NLP literature corpus by investigating the motivation, leveraged explainability methods, and corresponding challenges and solution concepts of its practical adoption. Our findings \textbf{have several implications for both research and practice}. For practitioners and industry people, documenting explainable NLP practices in the industry can provide valuable insights for other organizations seeking to implement explainable NLP solutions. These findings can serve as a reference for industry players aiming to enhance transparency and interpretability in their NLP applications. For researchers, understanding current industry practices enables assessing whether recent advancements in explainable NLP are being adopted in real-world applications. This alignment can help bridge some small gaps between academic research and practical implementation, ensuring that research efforts effectively address industry needs and challenges. We therefore encourage practitioners, researchers, and other interested readers to use our work as a foundation to complement and refine our results, potentially extending this research with their own insights and with follow-up studies to verify our presented findings and results. 

\textbf{Future work:} Further research could build on the specific findings of this study by extending it to include a larger and more diverse set of interviewees. This may involve focusing on specific stakeholder subgroups in high-stakes domains, such as patients or doctors in the medical field, to first identify their expectations and needs for effective explanations. These insights could then inform the development of domain- and user-tailored explainability methods and explanations. Furthermore, future research could explore the impact of emerging regulations, such as the EU AI Act, on the development of explainability methods and the design of explanations for NLP-based systems. 

\section*{Acknowledgment}
We thank the anonymous reviewers for their insightful and constructive feedback. This research has been supported by the German Federal Ministry of Education and Research (BMBF) grant 01IS23069 Software Campus 3.0 (TU München).

\bibliography{aaai25}

\end{document}